\documentclass[a4paper]{article}
\pdfoutput=1
\usepackage{INTERSPEECH2022}
\usepackage{latexsym}
\usepackage{amsmath}
\usepackage{url}
\usepackage{amssymb}
\usepackage{amsfonts}
\usepackage{graphicx}
\usepackage{tabularx}
\usepackage{multirow}
\usepackage{arydshln}
\usepackage{mathtools,nccmath}
\usepackage[T5]{fontenc}
\usepackage{enumitem}
\usepackage{todonotes}
\usepackage[most]{tcolorbox}
\usepackage[hidelinks]{hyperref}

\usepackage{helvet}  
\usepackage{courier}  
\usepackage{graphicx} 
\usepackage{algorithm}
\usepackage{algorithmic}
\usepackage{multirow}
\usepackage[T5]{fontenc}

\usepackage{amsmath}
\usepackage{url}
\usepackage{amssymb}
\usepackage{amsfonts}
\usepackage{graphicx}
\usepackage{tabularx}
\usepackage{multirow}
\usepackage{arydshln}
\usepackage{mathtools,nccmath}
\usepackage{listings}

\usepackage[T5]{fontenc}
\usepackage{enumitem}
\usepackage{todonotes}
\usepackage{cancel}
\usepackage[draft]{minted}

\makeatletter
\def\PYG@reset{\let\PYG@it=\relax \let\PYG@bf=\relax%
    \let\PYG@ul=\relax \let\PYG@tc=\relax%
    \let\PYG@bc=\relax \let\PYG@ff=\relax}
\def\PYG@tok#1{\csname PYG@tok@#1\endcsname}
\def\PYG@toks#1+{\ifx\relax#1\empty\else%
    \PYG@tok{#1}\expandafter\PYG@toks\fi}
\def\PYG@do#1{\PYG@bc{\PYG@tc{\PYG@ul{%
    \PYG@it{\PYG@bf{\PYG@ff{#1}}}}}}}
\def\PYG#1#2{\PYG@reset\PYG@toks#1+\relax+\PYG@do{#2}}

\expandafter\def\csname PYG@tok@w\endcsname{\def\PYG@tc##1{\textcolor[rgb]{0.73,0.73,0.73}{##1}}}
\expandafter\def\csname PYG@tok@c\endcsname{\let\PYG@it=\textit\def\PYG@tc##1{\textcolor[rgb]{0.25,0.50,0.50}{##1}}}
\expandafter\def\csname PYG@tok@cp\endcsname{\def\PYG@tc##1{\textcolor[rgb]{0.74,0.48,0.00}{##1}}}
\expandafter\def\csname PYG@tok@k\endcsname{\let\PYG@bf=\textbf\def\PYG@tc##1{\textcolor[rgb]{0.00,0.50,0.00}{##1}}}
\expandafter\def\csname PYG@tok@kp\endcsname{\def\PYG@tc##1{\textcolor[rgb]{0.00,0.50,0.00}{##1}}}
\expandafter\def\csname PYG@tok@kt\endcsname{\def\PYG@tc##1{\textcolor[rgb]{0.69,0.00,0.25}{##1}}}
\expandafter\def\csname PYG@tok@o\endcsname{\def\PYG@tc##1{\textcolor[rgb]{0.40,0.40,0.40}{##1}}}
\expandafter\def\csname PYG@tok@ow\endcsname{\let\PYG@bf=\textbf\def\PYG@tc##1{\textcolor[rgb]{0.67,0.13,1.00}{##1}}}
\expandafter\def\csname PYG@tok@nb\endcsname{\def\PYG@tc##1{\textcolor[rgb]{0.00,0.50,0.00}{##1}}}
\expandafter\def\csname PYG@tok@nf\endcsname{\def\PYG@tc##1{\textcolor[rgb]{0.00,0.00,1.00}{##1}}}
\expandafter\def\csname PYG@tok@nc\endcsname{\let\PYG@bf=\textbf\def\PYG@tc##1{\textcolor[rgb]{0.00,0.00,1.00}{##1}}}
\expandafter\def\csname PYG@tok@nn\endcsname{\let\PYG@bf=\textbf\def\PYG@tc##1{\textcolor[rgb]{0.00,0.00,1.00}{##1}}}
\expandafter\def\csname PYG@tok@ne\endcsname{\let\PYG@bf=\textbf\def\PYG@tc##1{\textcolor[rgb]{0.82,0.25,0.23}{##1}}}
\expandafter\def\csname PYG@tok@nv\endcsname{\def\PYG@tc##1{\textcolor[rgb]{0.10,0.09,0.49}{##1}}}
\expandafter\def\csname PYG@tok@no\endcsname{\def\PYG@tc##1{\textcolor[rgb]{0.53,0.00,0.00}{##1}}}
\expandafter\def\csname PYG@tok@nl\endcsname{\def\PYG@tc##1{\textcolor[rgb]{0.63,0.63,0.00}{##1}}}
\expandafter\def\csname PYG@tok@ni\endcsname{\let\PYG@bf=\textbf\def\PYG@tc##1{\textcolor[rgb]{0.60,0.60,0.60}{##1}}}
\expandafter\def\csname PYG@tok@na\endcsname{\def\PYG@tc##1{\textcolor[rgb]{0.49,0.56,0.16}{##1}}}
\expandafter\def\csname PYG@tok@nt\endcsname{\let\PYG@bf=\textbf\def\PYG@tc##1{\textcolor[rgb]{0.00,0.50,0.00}{##1}}}
\expandafter\def\csname PYG@tok@nd\endcsname{\def\PYG@tc##1{\textcolor[rgb]{0.67,0.13,1.00}{##1}}}
\expandafter\def\csname PYG@tok@s\endcsname{\def\PYG@tc##1{\textcolor[rgb]{0.73,0.13,0.13}{##1}}}
\expandafter\def\csname PYG@tok@sd\endcsname{\let\PYG@it=\textit\def\PYG@tc##1{\textcolor[rgb]{0.73,0.13,0.13}{##1}}}
\expandafter\def\csname PYG@tok@si\endcsname{\let\PYG@bf=\textbf\def\PYG@tc##1{\textcolor[rgb]{0.73,0.40,0.53}{##1}}}
\expandafter\def\csname PYG@tok@se\endcsname{\let\PYG@bf=\textbf\def\PYG@tc##1{\textcolor[rgb]{0.73,0.40,0.13}{##1}}}
\expandafter\def\csname PYG@tok@sr\endcsname{\def\PYG@tc##1{\textcolor[rgb]{0.73,0.40,0.53}{##1}}}
\expandafter\def\csname PYG@tok@ss\endcsname{\def\PYG@tc##1{\textcolor[rgb]{0.10,0.09,0.49}{##1}}}
\expandafter\def\csname PYG@tok@sx\endcsname{\def\PYG@tc##1{\textcolor[rgb]{0.00,0.50,0.00}{##1}}}
\expandafter\def\csname PYG@tok@m\endcsname{\def\PYG@tc##1{\textcolor[rgb]{0.40,0.40,0.40}{##1}}}
\expandafter\def\csname PYG@tok@gh\endcsname{\let\PYG@bf=\textbf\def\PYG@tc##1{\textcolor[rgb]{0.00,0.00,0.50}{##1}}}
\expandafter\def\csname PYG@tok@gu\endcsname{\let\PYG@bf=\textbf\def\PYG@tc##1{\textcolor[rgb]{0.50,0.00,0.50}{##1}}}
\expandafter\def\csname PYG@tok@gd\endcsname{\def\PYG@tc##1{\textcolor[rgb]{0.63,0.00,0.00}{##1}}}
\expandafter\def\csname PYG@tok@gi\endcsname{\def\PYG@tc##1{\textcolor[rgb]{0.00,0.63,0.00}{##1}}}
\expandafter\def\csname PYG@tok@gr\endcsname{\def\PYG@tc##1{\textcolor[rgb]{1.00,0.00,0.00}{##1}}}
\expandafter\def\csname PYG@tok@ge\endcsname{\let\PYG@it=\textit}
\expandafter\def\csname PYG@tok@gs\endcsname{\let\PYG@bf=\textbf}
\expandafter\def\csname PYG@tok@gp\endcsname{\let\PYG@bf=\textbf\def\PYG@tc##1{\textcolor[rgb]{0.00,0.00,0.50}{##1}}}
\expandafter\def\csname PYG@tok@ga\endcsname{\def\PYG@tc##1{\textcolor[rgb]{0.50,0.00,0.50}{##1}}}
\expandafter\def\csname PYG@tok@go\endcsname{\def\PYG@tc##1{\textcolor[rgb]{0.53,0.53,0.53}{##1}}}
\expandafter\def\csname PYG@tok@gt\endcsname{\def\PYG@tc##1{\textcolor[rgb]{0.00,0.27,0.87}{##1}}}
\expandafter\def\csname PYG@tok@err\endcsname{\def\PYG@bc##1{\setlength{\fboxsep}{0pt}\fcolorbox[rgb]{1.00,0.00,0.00}{1,1,1}{\strut ##1}}}
\expandafter\def\csname PYG@tok@kc\endcsname{\let\PYG@bf=\textbf\def\PYG@tc##1{\textcolor[rgb]{0.00,0.50,0.00}{##1}}}
\expandafter\def\csname PYG@tok@kd\endcsname{\let\PYG@bf=\textbf\def\PYG@tc##1{\textcolor[rgb]{0.00,0.50,0.00}{##1}}}
\expandafter\def\csname PYG@tok@kn\endcsname{\let\PYG@bf=\textbf\def\PYG@tc##1{\textcolor[rgb]{0.00,0.50,0.00}{##1}}}
\expandafter\def\csname PYG@tok@kr\endcsname{\let\PYG@bf=\textbf\def\PYG@tc##1{\textcolor[rgb]{0.00,0.50,0.00}{##1}}}
\expandafter\def\csname PYG@tok@bp\endcsname{\def\PYG@tc##1{\textcolor[rgb]{0.00,0.50,0.00}{##1}}}
\expandafter\def\csname PYG@tok@fm\endcsname{\def\PYG@tc##1{\textcolor[rgb]{0.00,0.00,1.00}{##1}}}
\expandafter\def\csname PYG@tok@vc\endcsname{\def\PYG@tc##1{\textcolor[rgb]{0.10,0.09,0.49}{##1}}}
\expandafter\def\csname PYG@tok@vg\endcsname{\def\PYG@tc##1{\textcolor[rgb]{0.10,0.09,0.49}{##1}}}
\expandafter\def\csname PYG@tok@vi\endcsname{\def\PYG@tc##1{\textcolor[rgb]{0.10,0.09,0.49}{##1}}}
\expandafter\def\csname PYG@tok@vm\endcsname{\def\PYG@tc##1{\textcolor[rgb]{0.10,0.09,0.49}{##1}}}
\expandafter\def\csname PYG@tok@sa\endcsname{\def\PYG@tc##1{\textcolor[rgb]{0.73,0.13,0.13}{##1}}}
\expandafter\def\csname PYG@tok@sb\endcsname{\def\PYG@tc##1{\textcolor[rgb]{0.73,0.13,0.13}{##1}}}
\expandafter\def\csname PYG@tok@sc\endcsname{\def\PYG@tc##1{\textcolor[rgb]{0.73,0.13,0.13}{##1}}}
\expandafter\def\csname PYG@tok@dl\endcsname{\def\PYG@tc##1{\textcolor[rgb]{0.73,0.13,0.13}{##1}}}
\expandafter\def\csname PYG@tok@s2\endcsname{\def\PYG@tc##1{\textcolor[rgb]{0.73,0.13,0.13}{##1}}}
\expandafter\def\csname PYG@tok@sh\endcsname{\def\PYG@tc##1{\textcolor[rgb]{0.73,0.13,0.13}{##1}}}
\expandafter\def\csname PYG@tok@s1\endcsname{\def\PYG@tc##1{\textcolor[rgb]{0.73,0.13,0.13}{##1}}}
\expandafter\def\csname PYG@tok@mb\endcsname{\def\PYG@tc##1{\textcolor[rgb]{0.40,0.40,0.40}{##1}}}
\expandafter\def\csname PYG@tok@mf\endcsname{\def\PYG@tc##1{\textcolor[rgb]{0.40,0.40,0.40}{##1}}}
\expandafter\def\csname PYG@tok@mh\endcsname{\def\PYG@tc##1{\textcolor[rgb]{0.40,0.40,0.40}{##1}}}
\expandafter\def\csname PYG@tok@mi\endcsname{\def\PYG@tc##1{\textcolor[rgb]{0.40,0.40,0.40}{##1}}}
\expandafter\def\csname PYG@tok@il\endcsname{\def\PYG@tc##1{\textcolor[rgb]{0.40,0.40,0.40}{##1}}}
\expandafter\def\csname PYG@tok@mo\endcsname{\def\PYG@tc##1{\textcolor[rgb]{0.40,0.40,0.40}{##1}}}
\expandafter\def\csname PYG@tok@ch\endcsname{\let\PYG@it=\textit\def\PYG@tc##1{\textcolor[rgb]{0.25,0.50,0.50}{##1}}}
\expandafter\def\csname PYG@tok@cm\endcsname{\let\PYG@it=\textit\def\PYG@tc##1{\textcolor[rgb]{0.25,0.50,0.50}{##1}}}
\expandafter\def\csname PYG@tok@cpf\endcsname{\let\PYG@it=\textit\def\PYG@tc##1{\textcolor[rgb]{0.25,0.50,0.50}{##1}}}
\expandafter\def\csname PYG@tok@c1\endcsname{\let\PYG@it=\textit\def\PYG@tc##1{\textcolor[rgb]{0.25,0.50,0.50}{##1}}}
\expandafter\def\csname PYG@tok@cs\endcsname{\let\PYG@it=\textit\def\PYG@tc##1{\textcolor[rgb]{0.25,0.50,0.50}{##1}}}

\def\PYGZus{\char`\_}

\def\PYGZsh{\char`\#}

\def\PYGZas{$\ast$}

\makeatother

\title{BARTpho: Pre-trained Sequence-to-Sequence Models for Vietnamese}

\name{Nguyen Luong Tran, Duong Minh Le, Dat Quoc Nguyen}
\address{VinAI Research, Hanoi, Vietnam }
\email{\{v.nguyentl12, v.duonglm1, v.datnq9\}@vinai.io}

\begin{document}
\maketitle

\begin{abstract}
We present BARTpho with two versions, BARTpho\textsubscript{syllable} and BARTpho\textsubscript{word}, which are the first public large-scale monolingual sequence-to-sequence models pre-trained for Vietnamese. BARTpho uses the ``large'' architecture and the pre-training scheme of the sequence-to-sequence denoising autoencoder BART, thus it is especially suitable for generative NLP tasks. We conduct experiments to compare our BARTpho with its competitor mBART on a downstream task of Vietnamese text summarization and show that: in  both  automatic  and  human  evaluations, BARTpho outperforms the strong baseline mBART and improves the state-of-the-art. We further evaluate and compare BARTpho and mBART on the Vietnamese capitalization and punctuation restoration tasks and also find that BARTpho is more effective than mBART on these two tasks. 
We publicly release BARTpho to facilitate future research and applications of generative Vietnamese NLP tasks. 
\end{abstract}


\medskip 
\noindent\textbf{Index Terms}: BARTpho; Sequence-to-Sequence; Vietnamese; Pre-trained models; Text summarization; Capitalization; Punctuation restoration. 

\section{Introduction}

The masked language model BERT \cite{devlin-etal-2019-bert} and its variants, pre-trained on large-scale corpora, help improve the state-of-the-art  (SOTA)  performances of various natural language understanding (NLU) tasks. However, due to a bidirectionality nature, it might be difficult to directly apply those pre-trained language models to natural language generation tasks \cite{wang2019bert}. Therefore, pre-trained sequence-to-sequence (seq2seq) models are proposed to handle this issue \cite{NEURIPS2019_c20bb2d9,lewis-etal-2020-bart,pmlr-v119-zhang20ae,JMLR:v21:20-074,qi-etal-2020-prophetnet,byt5}. The success of these pre-trained seq2seq models has largely been limited to the English language. 
From a societal,  cultural, linguistic, cognitive and machine learning perspective \cite{donlpotherlanguages}, it is worth investigating pre-trained seq2seq models for languages other than English. 
For other languages, one could employ existing pre-trained multilingual seq2seq models \cite{tacl_a_00343,xue-etal-2021-mt5,qi-etal-2021-prophetnet} or retrain  language-specific models using the proposed seq2seq  architectures \cite{eddine2020barthez,bartchinese}. Note that retraining a language-specific model might be preferable as dedicated language-specific models still outperform multilingual ones \cite{nguyen-tuan-nguyen-2020-phobert}.

Regarding Vietnamese, to the best of our knowledge, there is not an existing monolingual seq2seq model pre-trained for Vietnamese. In addition, another concern is that all publicly available pre-trained multilingual seq2seq models are not aware of the linguistic characteristic difference between Vietnamese syllables and word tokens.  
This comes from the fact that when written in Vietnamese, in addition to marking word boundaries, the white space is also used to separate syllables that constitute words.\footnote{Note that 85\% of Vietnamese word types are composed of at least two syllables \cite{DinhQuangThang2008}.} For example, a 7-syllable written text ``Chúng tôi là những nghiên cứu viên''\textsubscript{We are researchers} forms a 4-word text ``Chúng\_tôi\textsubscript{We} là\textsubscript{are} những nghiên\_cứu\_viên\textsubscript{reseacher}''.  Without applying a Vietnamese word segmenter, those pre-trained multilingual seq2seq models directly apply Byte-Pair encoding models \cite{sennrich-etal-2016-neural,kudo-richardson-2018-sentencepiece} to the syllable-level Vietnamese pre-training data. 
Therefore, it is worth investigating the influence of word segmentation on seq2seq pre-training for Vietnamese. 

In this paper, we introduce BARTpho with two versions---BARTpho\textsubscript{syllable} and BARTpho\textsubscript{word}---the first large-scale monolingual seq2seq models pre-trained for Vietnamese, which are based on the seq2seq denoising autoencoder BART \cite{lewis-etal-2020-bart}. 
The difference between our two BARTpho versions is that they take different types of input texts: a syllable level for BARTpho\textsubscript{syllable} vs. a word level for BARTpho\textsubscript{word}. We compare BARTpho with mBART \cite{tacl_a_00343}---a multilingual variant of BART---on a downstream task of Vietnamese text summarization, and we find that our BARTpho models outperform mBART in both automatic and human evaluations, and help produce a new SOTA performance, thus showing the effectiveness of large-scale monolingual seq2seq pre-training for Vietnamese. We also evaluate and compare BARTpho and mBART on the Vietnamese capitalization and punctuation restoration tasks and find that BARTpho helps produce better performance results than mBART. In all three evaluation tasks, we find that BARTpho\textsubscript{word} does better than BARTpho\textsubscript{syllable}, showing the positive influence of Vietnamese word segmentation towards seq2seq pre-training. 

We publicly release our BARTpho models at \url{https://github.com/VinAIResearch/BARTpho},  which can be used with popular libraries  \texttt{fairseq}  \cite{ott2019fairseq} and  \texttt{transformers} \cite{wolf-etal-2020-transformers}. We hope that our BARTpho can serve as a strong baseline for future research and applications of generative natural language processing (NLP) tasks for Vietnamese.

\begin{figure*}[!t]
\begin{Verbatim}[commandchars=\\\{\}]
\PYG{k+kn}{from} {transformers} \PYG{k+kn}{import} \PYG{l+s+s2}{AutoModel}, \PYG{l+s+s2}{AutoTokenizer}

\PYG{c+c1}{\PYGZsh{} BARTpho\textsubscript{syllable}}
tokenizer = \PYG{l+s+s2}{AutoTokenizer}.\PYG{ga}{from\PYGZus{}pretrained}(\PYG{vi}{"vinai/bartpho-syllable"})
bartpho\PYGZus{}syllable = \PYG{l+s+s2}{AutoModel}.\PYG{ga}{from\PYGZus{}pretrained}(\PYG{vi}{"vinai/bartpho-syllable"})
input\PYGZus{}text = \PYG{vi}{'Chúng tôi là những nghiên cứu viên'}
input\PYGZus{}ids = \PYG{ga}{tokenizer}(input\PYGZus{}text, return\PYGZus{}tensors=\PYG{vi}{'pt'})
features = \PYG{ga}{bartpho\PYGZus{}syllable}(\PYGZas{}\PYGZas{}input\PYGZus{}ids)

\PYG{c+c1}{\PYGZsh{} BARTpho\textsubscript{word}}
tokenizer = \PYG{l+s+s2}{AutoTokenizer}.\PYG{ga}{from\PYGZus{}pretrained}(\PYG{vi}{"vinai/bartpho-word"})
bartpho\PYGZus{}word = \PYG{l+s+s2}{AutoModel}.\PYG{ga}{from\PYGZus{}pretrained}(\PYG{vi}{"vinai/bartpho-word"})
input\PYGZus{}text = \PYG{vi}{'Chúng\_tôi là những nghiên\_cứu\_viên'}
input\PYGZus{}ids = \PYG{ga}{tokenizer}(input\PYGZus{}text, return\PYGZus{}tensors=\PYG{vi}{'pt'})
features = \PYG{ga}{bartpho\PYGZus{}word}(\PYGZas{}\PYGZas{}input\PYGZus{}ids)
\end{Verbatim}
\caption{An example code using BARTpho for feature extraction with \texttt{transformers} in Python. Here, a 7-syllable text sequence ``Chúng tôi là những nghiên cứu viên''\textsubscript{We are researchers} forms a 4-word sequence ``Chúng\_tôi\textsubscript{We} là\textsubscript{are} những nghiên\_cứu\_viên\textsubscript{reseacher}''.}
\label{fig:code}
\end{figure*}

\section{Related work}

PhoBERT \cite{nguyen-tuan-nguyen-2020-phobert} is the first public large-scale monolingual language model pre-trained for Vietnamese, which helps obtain  state-of-the-art performances on various downstream Vietnamese NLP/NLU tasks \cite{PhoNER_COVID19, phonlp, JointIDSF,aspectvnamese,vitext2sql}. PhoBERT is pre-trained on a 20GB word-level corpus of Vietnamese texts, using the RoBERTa pre-training approach \cite{RoBERTa} that optimizes BERT for more robust performance. Following PhoBERT, there are also public monolingual language models for Vietnamese such as viBERT and vELECTRA \cite{bui-etal-2020-improving}, which are based on BERT and ELECTRA pre-training approaches \cite{devlin-etal-2019-bert,clark2020electra} and pre-trained on syllable-level Vietnamese text corpora. Following  Rothe et al. \cite{rothe-etal-2020-leveraging} who leverage pre-trained language model checkpoints for sequence generation tasks, Nguyen et al. \cite{vietsum} conduct an empirical study and show that PhoBERT helps produce better performance results than viBERT for a downstream task of Vietnamese abstractive summarization. 

Our BARTpho is based on BART. We employ BART because it helps produce the strongest performances  on  downstream tasks in comparison to other pre-trained seq2seq models under a comparable setting in terms  of the relatively equal numbers of model parameters and pre-training data sizes  \cite{lewis-etal-2020-bart,JMLR:v21:20-074,qi-etal-2020-prophetnet}. BART is also used to pre-train monolingual models for other languages such as French \cite{eddine2020barthez} and Chinese \cite{bartchinese}.

\section{Our BARTpho}
This section describes the architecture, the pre-training data and the optimization setup, that we use for BARTpho. 

\subsection{Architecture}
Both BARTpho\textsubscript{syllable} and BARTpho\textsubscript{word} use the ``large'' architecture with 12 encoder and decoder layers and pre-training scheme of BART  \cite{lewis-etal-2020-bart}. In particular, pre-training BART has two stages: (i) corrupting the input text with an arbitrary noising function, and (ii) learning to reconstruct the original text, i.e. optimizing the cross-entropy between its decoder’s output and the original text. Here, BART uses the standard architecture Transformer \cite{NIPS2017_3f5ee243}, but employing the GeLU activation function \cite{hendrycks2016gelu} rather than ReLU and performing parameter initialization from $\mathcal{N}$(0, 0.02). 
Following BART \cite{lewis-etal-2020-bart}, we employ two types of noise in the noising function, including text infilling and sentence permutation. For text infilling, we sample a number of text spans with their lengths drawn from a Poisson distribution ($\lambda$ = 3.5) and replace each span with a single special $<$mask$>$ token. For sentence permutation,
consecutive sentences are grouped to generate sentence blocks of 512 tokens, and sentences in each block are then shuffled in random order.  Following mBART \cite{tacl_a_00343}, we also add a layer-normalization layer on top of both the encoder and decoder.

\subsection{Pre-training data}

For BARTpho\textsubscript{word}, we employ the PhoBERT pre-training corpus \cite{nguyen-tuan-nguyen-2020-phobert}, that contains 20GB of uncompressed texts (about 145M automatically word-segmented sentences). 
In addition, we also reuse the PhoBERT's tokenizer that applies a vocabulary of 64K subword types and BPE \cite{sennrich-etal-2016-neural} to segment those word-segmented sentences with subword units. BARTpho\textsubscript{word} has about 420M parameters. 
Pre-training data for BARTpho\textsubscript{syllable} is a detokenized variant of the PhoBERT pre-training corpus (i.e. about 4B syllable tokens). We employ the pre-trained SentencePiece model \cite{kudo-richardson-2018-sentencepiece} from XLM-RoBERTa \cite{conneau2019unsupervised}, used in mBART  \cite{tacl_a_00343}, to segment sentences with sub-syllable units and select a vocabulary of the top 40K most frequent types. BARTpho\textsubscript{syllable} has about 396M parameters. 

\subsection{Optimization}

We utilize the BART implementation with the denoising task from \texttt{fairseq} \cite{ott2019fairseq}. We use Adam \cite{KingmaB14} for optimization, and use a batch size of 512 sequence blocks across 8 A100 GPUs (40GB each) and a peak learning rate of 0.0001. Note that we initialize parameter weights of BARTpho\textsubscript{syllable} by those from mBART. For each BARTpho model, we run for 15 training epochs  in about 6 days (here, the learning rate is warmed up for 1.5 epochs).

\subsection{Usage example}

Figure \ref{fig:code} presents a basic usage of our pre-trained BARTpho models for feature extraction with  \texttt{transformers}  to show its potential use for other downstream tasks.\footnote{\url{https://huggingface.co/docs/transformers/model_doc/bartpho}} More usage examples of BARTpho with both \texttt{fairseq} and  \texttt{transformers} can be found at the BARTpho's GitHub repository.

\begin{table*}[!t]

\centering
\caption{Detokenized and case-sensitive ROUGE scores (in \%) w.r.t. duplicate article removal. R-1, R-2 and R-L abbreviate ROUGE-1, ROUGE-2 and ROUGE-L, respectively. Every score difference between mBART and each BARTpho version is statistically significant with p-value $<$ 0.05.}
\def\arraystretch{1.1}
\begin{tabular}{l|c|c|c|c|c|c|c}
\hline
\multirow{2}{*}{\textbf{Model}} & \multicolumn{3}{c|}{\textbf{Validation set}} & \multicolumn{4}{c}{\textbf{Test set}} \\
\cline{2-8}
& \textbf{R-1} & \textbf{R-2}   & \textbf{R-L} &  \textbf{R-1} & \textbf{R-2}   & \textbf{R-L}  & \textbf{Human} \\
\hline
mBART & 60.06 & 28.69 & 38.85 & 60.03 & 28.51 & 38.74 & 21/100\\
BARTpho\textsubscript{syllable} & \underline{60.29} & \underline{29.07} & \underline{39.02} & \underline{60.41} & \underline{29.20} & \underline{39.22} & \underline{37/100}\\
BARTpho\textsubscript{word} & \textbf{60.55} & \textbf{29.89} & \textbf{39.73} & \textbf{60.51} & \textbf{29.65} & \textbf{39.75} & \textbf{42/100}\\
\hline
\end{tabular}

\label{tab:results}
\end{table*}

\begin{table*}[!t]
\centering
\caption{ROUGE scores (in \%) w.r.t. the original dataset setting (i.e.  {without} duplicate article removal). \textbf{[$\star$]} denotes the best performing model among different  models experimented from \cite{9023886}, and \textbf{[$\ast$]} denotes scores reported in \cite{vietsum}.}
\def\arraystretch{1.1}
\begin{tabular}{l|c|c|c|c|c|c}
\hline
\multirow{2}{*}{\textbf{Model}}  & \multicolumn{3}{c|}{\textbf{Original validation set}} & \multicolumn{3}{c}{\textbf{Original test set}}\\
\cline{2-7}
& \textbf{R-1} & \textbf{R-2}   & \textbf{R-L} & \textbf{R-1} & \textbf{R-2}   & \textbf{R-L} \\
\hline
fastAbs \textbf{[$\star$]} & \_ & \_ & \_ & 54.52 & 23.01 & 37.64 \\
viBERT2viBERT \textbf{[$\ast$]} & \_ & \_ & \_ & 59.75 & 27.29 & 36.79 \\
PhoBERT2PhoBERT \textbf{[$\ast$]} & \_ & \_ & \_ & {60.37} & {29.12} & {39.44} \\
mT5 \textbf{[$\ast$]} & \_ & \_ & \_ & 58.05 & 26.76 & 37.38\\
\hline
\hline 
mBART & 60.39 & 29.19 & 39.18 & 60.35 & 29.13 & 39.21\\
BARTpho\textsubscript{syllable} & \underline{60.89} & \underline{29.98} & \underline{39.59} & \underline{60.88} & \underline{29.90} & \underline{39.64}\\
BARTpho\textsubscript{word} & \textbf{61.10} & \textbf{30.34} & \textbf{40.05} & \textbf{61.14} & \textbf{30.31} & \textbf{40.15}\\
\hline
\end{tabular}
\label{tab:originaltest}
\end{table*}

\section{Experiments}

\subsection{Text summarization}\label{sec:sum}

We evaluate and compare the performance of BARTpho with the strong baseline mBART on a downstream generative task of Vietnamese text summarization. 
Here, mBART is pre-trained on a Common Crawl dataset of 25 languages, which includes 137 GB of syllable-level Vietnamese texts. 

\subsubsection{Experimental setup}\label{ssec:setup}

We employ the single-document summarization dataset VNDS  \cite{9023886}, consisting of 150704 news articles each including a news abstract (i.e. gold summary) and body content (i.e. input text). 
In particular, 105418, 22642 and 22644 articles are used for training, validation and test, respectively. However, we find that there are duplicate articles in this dataset. Therefore, we filter the duplicates, resulting in 99134, 22184 and 22498 articles for training, validation and test, respectively.\footnote{Firstly, we remove duplicates inside each of the training, validation and test sets. Secondly, if an article appears in both training and validation/test sets, then the article is filtered out of the training set. Lastly, if an article appears in both validation and test sets, then the article is filtered out of the validation set.} When fine-tuning BARTpho\textsubscript{syllable} and mBART, we use a detokenized version of the filtered dataset, while its automatically word-segmented version is used for fine-tuning BARTpho\textsubscript{word}. 

We formulate this task as a monolingual translation problem and fine-tune our BARTpho and the baseline mBART using the same hyper-parameter tuning strategy. We fix the maximum number of tokens in a batch at 4096. We use Adam and run for 20 training epochs.  We also perform grid search to select the Adam initial learning rate from \{1e-5, 2e-5, 3e-5, 5e-5\}. We employ beam search with a beam size of 4 for decoding. We evaluate each model 4 times in every epoch. We select the model checkpoint that produces the highest ROUGE-L score \cite{lin-2004-rouge} on the validation set, and we then apply the selected one to the test set. Note that we compute the detokenized and case-sensitive ROUGE scores for all models (here, we detokenize the fine-tuned BARTpho\textsubscript{word}'s output before computing the scores).

\subsubsection{Main results}

\begin{table*}[!t]
\centering
\caption{Capitalization and punctuation restoration  F\textsubscript{1} scores (in \%) on the test set. Due to the space limit, we do not include scores on the validation set. Note that we also observe similar findings on the validation set. }

\def\arraystretch{1.1}
\begin{tabular}{l|c|c|c|c|c}
\hline
\multirow{2}{*}{\textbf{Model}} & \multirow{2}{*}{\textbf{Capitalization}} & \multicolumn{4}{c}{\textbf{Punctuation restoration}} \\
\cline{3-6}
& & \textbf{Comma} & \textbf{Period}  
& \textbf{Question} & \textbf{Overall}
\\
\hline
mBART & 91.28 &  67.26 & \textbf{92.19} & 85.71  & 78.71  \\
BARTpho\textsubscript{syllable} & \underline{91.98} & \underline{67.95} & 91.79 & \textbf{88.15} & \underline{79.09}  \\
BARTpho\textsubscript{word} & \textbf{92.41} &  \textbf{68.39} & \underline{92.05} & \underline{87.82}  & \textbf{79.29}   \\
\hline
\end{tabular}

\label{tab:punct_results}
\end{table*}

Table \ref{tab:results} presents our obtained ROUGE scores on the validation and test sets for the baseline mBART and our two BARTpho versions w.r.t. the setting of duplicate article removal. Clearly, both BARTpho versions achieve  significantly better  ROUGE scores than mBART on both validation and test sets.

We also conduct a human-based manual comparison between the outputs produced by the baseline mBART and our two BARTpho versions. In particular, we randomly sample 100
input text examples from the test set; and for each input example, we anonymously shuffle the summary outputs from three fine-tuned models (here, each input sampled example satisfies that any two out of three summary outputs are not exactly the same).  We then ask two external Vietnamese annotators to choose which summary they think is the best. We obtain a Cohen's kappa coefficient at 0.61 for the inter-annotator agreement between the two annotators. Our second co-author then hosts and participates in a discussion session with the two annotators to resolve annotation conflicts (here, he does not know which model produces which summary). Table \ref{tab:results} shows final scores where our BARTpho obtains a better human evaluation result than mBART.

For comparison with previously published results \cite{9023886,vietsum}, we also fine-tune our BARTpho models and baseline mBART on the original training set  (i.e. without duplicate article removal),\footnote{This is not a proper experimental setup because of data leakage, e.g. 1466 training articles appear in the test set.} using the same hyper-parameter tuning strategy as presented in Section \ref{ssec:setup}. We report ROUGE scores on the original test set in Table \ref{tab:originaltest}. 
The previous best model from experiments in \cite{9023886,vietsum} is PhoBERT2PhoBERT with a ROUGE-L score at 39.44. This score is 0.2 and 0.7 points lower than those of BARTpho\textsubscript{syllable} and BARTpho\textsubscript{word}, respectively.  
Tables  \ref{tab:results} and \ref{tab:originaltest} show that BARTpho helps attain a new SOTA performance for this task. 

Our automatic and human evaluation results from tables  \ref{tab:results} and \ref{tab:originaltest} demonstrate the effectiveness of large-scale BART-based monolingual seq2seq models for Vietnamese. Note that mBART uses 137 / 20 $\approx$ 7 times bigger Vietnamese pre-training data than BARTpho. In addition, the multilingual seq2seq mT5 \cite{xue-etal-2021-mt5} is pre-trained on the multilingual dataset mC4 that includes 79M Common Crawl Vietnamese pages consisting of 116B syllable tokens, i.e.  mT5 uses 116 / 4 = 29 times bigger Vietnamese pre-training data than BARTpho. However, BARTpho surpasses both mBART and mT5, reconfirming that the dedicated language-specific model still performs better than the multilingual one \cite{nguyen-tuan-nguyen-2020-phobert}. Tables \ref{tab:results} and \ref{tab:originaltest} also show that BARTpho\textsubscript{word} outperforms BARTpho\textsubscript{syllable}, thus demonstrating the positive influence of word segmentation for seq2seq pre-training and fine-tuning in Vietnamese.

\subsection{Capitalization and punctuation restoration}

Most Automatic Speech Recognition (ASR) systems generate text transcripts without information about capitalization and punctuation, which limits the readability of the transcripts. 
In addition, using these lowercasing and non-punctuation types of ASR transcripts as input to downstream task models, e.g. named entity recognition, machine translation and the like, might also cause performance degradation \cite{tundik18_interspeech} because the downstream task models are usually trained on well-formatted text datasets. Thus, capitalization and punctuation restoration are important steps in ASR transcript post-processing. An example enriching ASR transcripts with capitalization and punctuation restoration is 
as follows: 

\medskip

\begin{tcolorbox}[title=A transcript]
chuỗi nhà hàng này gần đây đã phải đóng cửa một loạt các chi nhánh theo sở kế hoạch và đầu tư hà nội và thành phố hồ chí minh golden gate đã đóng cửa bảy chi nhánh vào cuối năm 2015
\end{tcolorbox}

\begin{tcolorbox}[title=The transcript enriched with capitalization and punctuation restoration \& its English translation]
Chuỗi nhà hàng này gần đây đã phải đóng cửa một loạt các chi nhánh. Theo Sở Kế hoạch và Đầu tư Hà Nội và Thành phố Hồ Chí Minh, Golden Gate đã đóng cửa bảy chi nhánh vào cuối năm 2015. \\
The chain has recently had to shut down a series of branches. According to the Hanoi and Ho Chi Minh City Planning and Investment Departments, the Golden Gate closed seven branches by the end of 2015.
\end{tcolorbox}

Capitalization and punctuation restoration models generally fall into two main categories of approaches: sequence tagging \cite{NguyenNTDM20,huang21g_interspeech, shi21_interspeech} and sequence-to-sequence \cite{NGUYEN20212020BDP0005,capu_binhnguyen}. In this investigation, we follow the sequence-to-sequence approach to evaluate and compare our BARTpho and mBART on the Vietnamese capitalization and punctuation restoration tasks. The models take lowercase, unpunctuated texts as input and produce true case, punctuated texts as output.

\subsubsection{Experimental setup}

Due to the lack of benchmark datasets for Vietnamese capitalization and punctuation restoration, we generate a dataset automatically by leveraging the PhoST dataset  \cite{phost} that contains 327370, 1933, and 1976  Vietnamese examples for training, validation and test, respectively. 
We convert those examples into a lowercase form and remove all punctuations to simulate the ASR transcript output. Here, the standard formats for numbers and currencies are retained. 
Following previous work \cite{NguyenNTDM20,NGUYEN20212020BDP0005}, we only consider three types of punctuation marks, which are Comma (includes commas, colons, and dashes), Period (includes full stops, exclamation marks, and semicolons), and Question (only question mark). 

We use the same fine-tuning procedure that we use for the  summarization task as presented in Section \ref{ssec:setup}. Here, for fine-tuning BARTpho\textsubscript{word}, we perform an automatic Vietnamese word segmentation on the data using RDRSegmenter \cite{rdrsegmenter} from the VnCoreNLP toolkit \cite{vu-etal-2018-vncorenlp}. We detokenize the
fine-tuned BARTpho\textsubscript{word}’s output before computing scores. Note that we select the model checkpoint that produces the lowest loss on the validation set and we apply the selected one to the test set.

\subsubsection{Main results}
Table \ref{tab:punct_results} presents the results obtained by our BARTpho and mBART  on the capitalization task. We find that our BARTpho performs better than mBART. In particular, BARTpho\textsubscript{word}  and BARTpho\textsubscript{syllable}  obtain  1.1\% and 0.7\% absolute higher F\textsubscript{1} scores than mBART, respectively.

Table \ref{tab:punct_results} also shows the obtained results of our BARTpho and mBART on the punctuation restoration task. Both BARTpho versions outperform mBART on the Comma and Question types, and the performance gap is substantial w.r.t. the latter mark. Furthermore, mBART does better than BARTpho on the Period mark, however, the performance gaps are small, i.e.  mBART produces 0.14\% and 0.4\% higher scores than BARTpho\textsubscript{word} and BARTpho\textsubscript{syllable}, respectively. Overall, our BARTpho still  outperforms mBART, where BARTpho\textsubscript{word} obtains the highest Overall F\textsubscript{1} score.

\section{Conclusion}

In this paper, we have presented BARTpho\textsubscript{syllable} and BARTpho\textsubscript{word}---the first pre-trained and large-scale monolingual seq2seq models for Vietnamese. We demonstrate the usefulness of our BARTpho by showing that BARTpho performs better than its competitor mBART and helps produce the SOTA performance for the downstream generative task of Vietnamese text summarization. We also show that BARTpho is more effective than mBART on the Vietnamese capitalization and punctuation restoration tasks. We hope that our public BARTpho models can foster future research and applications of generative Vietnamese NLP tasks.

\bibliography{refs}

\begin{thebibliography}{10}
\providecommand{\url}[1]{#1}
\csname url@samestyle\endcsname
\providecommand{\newblock}{\relax}
\providecommand{\bibinfo}[2]{#2}
\providecommand{\BIBentrySTDinterwordspacing}{\spaceskip=0pt\relax}
\providecommand{\BIBentryALTinterwordstretchfactor}{4}
\providecommand{\BIBentryALTinterwordspacing}{\spaceskip=\fontdimen2\font plus
\BIBentryALTinterwordstretchfactor\fontdimen3\font minus
  \fontdimen4\font\relax}
\providecommand{\BIBforeignlanguage}[2]{{%
\expandafter\ifx\csname l@#1\endcsname\relax
\typeout{** WARNING: IEEEtran.bst: No hyphenation pattern has been}%
\typeout{** loaded for the language `#1'. Using the pattern for}%
\typeout{** the default language instead.}%
\else
\language=\csname l@#1\endcsname
\fi
#2}}
\providecommand{\BIBdecl}{\relax}
\BIBdecl

\bibitem{devlin-etal-2019-bert}
J.~Devlin, M.-W. Chang, K.~Lee, and K.~Toutanova, ``{BERT}: Pre-training of
  deep bidirectional transformers for language understanding,'' in
  \emph{NAACL}, 2019.

\bibitem{wang2019bert}
A.~Wang and K.~Cho, ``{BERT has a Mouth, and It Must Speak: BERT as a Markov
  Random Field Language Model},'' \emph{arXiv preprint}, vol. arXiv:1902.04094,
  2019.

\bibitem{NEURIPS2019_c20bb2d9}
L.~Dong, N.~Yang, W.~Wang, F.~Wei, X.~Liu, Y.~Wang, J.~Gao, M.~Zhou, and H.-W.
  Hon, ``{Unified Language Model Pre-training for Natural Language
  Understanding and Generation},'' in \emph{NeurIPS}, vol.~32, 2019.

\bibitem{lewis-etal-2020-bart}
M.~Lewis, Y.~Liu, N.~Goyal, M.~Ghazvininejad, A.~Mohamed, O.~Levy, V.~Stoyanov,
  and L.~Zettlemoyer, ``{{BART}: Denoising Sequence-to-Sequence Pre-training
  for Natural Language Generation, Translation, and Comprehension},'' in
  \emph{ACL}, 2020.

\bibitem{pmlr-v119-zhang20ae}
J.~Zhang, Y.~Zhao, M.~Saleh, and P.~Liu, ``{PEGASUS}: Pre-training with
  extracted gap-sentences for abstractive summarization,'' in \emph{ICML},
  2020.

\bibitem{JMLR:v21:20-074}
C.~Raffel, N.~Shazeer, A.~Roberts, K.~Lee, S.~Narang, M.~Matena, Y.~Zhou,
  W.~Li, and P.~J. Liu, ``{Exploring the Limits of Transfer Learning with a
  Unified Text-to-Text Transformer},'' \emph{Journal of Machine Learning
  Research}, vol.~21, no. 140, 2020.

\bibitem{qi-etal-2020-prophetnet}
W.~Qi, Y.~Yan, Y.~Gong, D.~Liu, N.~Duan, J.~Chen, R.~Zhang, and M.~Zhou,
  ``{{P}rophet{N}et: Predicting Future N-gram for
  Sequence-to-{S}equence{P}re-training},'' in \emph{Findings of EMNLP}, 2020.

\bibitem{byt5}
L.~Xue, A.~Barua, N.~Constant, R.~Al-Rfou, S.~Narang, M.~Kale, A.~Roberts, and
  C.~Raffel, ``{ByT5: Towards a token-free future with pre-trained byte-to-byte
  models},'' \emph{arXiv preprint}, vol. arXiv:2105.13626, 2021.

\bibitem{donlpotherlanguages}
S.~Ruder, ``{Why You Should Do NLP Beyond English},''
  https://ruder.io/nlp-beyond-english/, 2020.

\bibitem{tacl_a_00343}
Y.~Liu, J.~Gu, N.~Goyal, X.~Li, S.~Edunov, M.~Ghazvininejad, M.~Lewis, and
  L.~Zettlemoyer, ``{Multilingual Denoising Pre-training for Neural Machine
  Translation},'' \emph{Transactions of the ACL}, vol.~8, 2020.

\bibitem{xue-etal-2021-mt5}
L.~Xue, N.~Constant, A.~Roberts, M.~Kale, R.~Al-Rfou, A.~Siddhant, A.~Barua,
  and C.~Raffel, ``{m{T}5: A Massively Multilingual Pre-trained Text-to-Text
  Transformer},'' in \emph{NAACL}, 2021.

\bibitem{qi-etal-2021-prophetnet}
W.~Qi, Y.~Gong, Y.~Yan, C.~Xu, B.~Yao, B.~Zhou, B.~Cheng, D.~Jiang, J.~Chen,
  R.~Zhang, H.~Li, and N.~Duan, ``{{P}rophet{N}et-{X}: Large-Scale Pre-training
  Models for {E}nglish, {C}hinese, Multi-lingual, Dialog, and Code
  Generation},'' in \emph{ACL: System Demonstrations}, 2021.

\bibitem{eddine2020barthez}
M.~K. Eddine, A.~J.-P. Tixier, and M.~Vazirgiannis, ``{BARThez: a Skilled
  Pretrained French Sequence-to-Sequence Model},'' \emph{arXiv preprint}, vol.
  arXiv:2010.12321, 2020.

\bibitem{bartchinese}
Y.~Shao, Z.~Geng, Y.~Liu, J.~Dai, F.~Yang, L.~Zhe, H.~Bao, and X.~Qiu, ``{CPT:
  A Pre-Trained Unbalanced Transformer for Both Chinese Language Understanding
  and Generation},'' \emph{arXiv preprint}, vol. arXiv:2109.05729, 2021.

\bibitem{nguyen-tuan-nguyen-2020-phobert}
D.~Q. Nguyen and A.~T. Nguyen, ``{P}ho{BERT}: Pre-trained language models for
  {V}ietnamese,'' in \emph{Findings of EMNLP}, 2020.

\bibitem{DinhQuangThang2008}
D.~Q. Thang, L.~H. Phuong, N.~T.~M. Huyen, N.~C. Tu, M.~Rossignol, and V.~X.
  Luong, ``{Word segmentation of Vietnamese texts: a comparison of
  approaches},'' in \emph{LREC}, 2008.

\bibitem{sennrich-etal-2016-neural}
R.~Sennrich, B.~Haddow, and A.~Birch, ``{Neural Machine Translation of Rare
  Words with Subword Units},'' in \emph{ACL}, 2016.

\bibitem{kudo-richardson-2018-sentencepiece}
T.~Kudo and J.~Richardson, ``{{S}entence{P}iece: A simple and language
  independent subword tokenizer and detokenizer for Neural Text Processing},''
  in \emph{EMNLP: System Demonstrations}, 2018.

\bibitem{ott2019fairseq}
M.~Ott, S.~Edunov, A.~Baevski, A.~Fan, S.~Gross, N.~Ng, D.~Grangier, and
  M.~Auli, ``{fairseq: A Fast, Extensible Toolkit for Sequence Modeling},'' in
  \emph{NAACL-HLT 2019: Demonstrations}, 2019.

\bibitem{wolf-etal-2020-transformers}
T.~Wolf, L.~Debut \emph{et~al.}, ``{Transformers: State-of-the-Art Natural
  Language Processing},'' in \emph{EMNLP 2020: System Demonstrations}, 2020.

\bibitem{PhoNER_COVID19}
T.~H. Truong, M.~H. Dao, and D.~Q. Nguyen, ``{COVID-19 Named Entity Recognition
  for Vietnamese},'' in \emph{NAACL-HLT}, 2021.

\bibitem{phonlp}
L.~T. Nguyen and D.~Q. Nguyen, ``{PhoNLP: A joint multi-task learning model for
  Vietnamese part-of-speech tagging, named entity recognition and dependency
  parsing},'' in \emph{NAACL: Demonstrations}, 2021.

\bibitem{JointIDSF}
M.~H. Dao, T.~H. Truong, and D.~Q. Nguyen, ``{Intent Detection and Slot Filling
  for Vietnamese},'' in \emph{INTERSPEECH}, 2021.

\bibitem{aspectvnamese}
D.~V. Thin, L.~S. Le, V.~X. Hoang, and N.~L.-T. Nguyen, ``{Investigating
  Monolingual and Multilingual BERT Models for Vietnamese Aspect Category
  Detection},'' \emph{arXiv preprint}, vol. arXiv:2103.09519, 2021.

\bibitem{vitext2sql}
A.~T. Nguyen, M.~H. Dao, and D.~Q. Nguyen, ``{A Pilot Study of Text-to-SQL
  Semantic Parsing for Vietnamese},'' in \emph{Findings of EMNLP}, 2020.

\bibitem{RoBERTa}
Y.~Liu, M.~Ott, N.~Goyal, J.~Du, M.~Joshi, D.~Chen, O.~Levy, M.~Lewis,
  L.~Zettlemoyer, and V.~Stoyanov, ``{RoBERTa: {A} Robustly Optimized {BERT}
  Pretraining Approach},'' \emph{arXiv preprint}, vol. arXiv:1907.11692, 2019.

\bibitem{bui-etal-2020-improving}
T.~V. Bui, T.~O. Tran, and P.~Le-Hong, ``{Improving Sequence Tagging for
  {V}ietnamese Text using Transformer-based Neural Models},'' in \emph{PACLIC},
  2020.

\bibitem{clark2020electra}
K.~Clark, M.-T. Luong, Q.~V. Le, and C.~D. Manning, ``{ELECTRA}: Pre-training
  text encoders as discriminators rather than generators,'' in \emph{ICLR},
  2020.

\bibitem{rothe-etal-2020-leveraging}
S.~Rothe, S.~Narayan, and A.~Severyn, ``{Leveraging Pre-trained Checkpoints for
  Sequence Generation Tasks},'' \emph{Transactions of the ACL}, vol.~8, 2020.

\bibitem{vietsum}
H.~Nguyen, L.~Phan, J.~Anibal, A.~Peltekian, and H.~Tran, ``{VieSum: How Robust
  Are Transformer-based Models on Vietnamese Summarization?}'' \emph{arXiv
  preprint}, vol. arXiv:2110.04257v1, 2021.

\bibitem{NIPS2017_3f5ee243}
A.~Vaswani, N.~Shazeer, N.~Parmar, J.~Uszkoreit, L.~Jones, A.~N. Gomez, L.~u.
  Kaiser, and I.~Polosukhin, ``{Attention is All you Need},'' in \emph{NIPS},
  vol.~30, 2017.

\bibitem{hendrycks2016gelu}
D.~Hendrycks and K.~Gimpel, ``{Gaussian Error Linear Units (GELUs)},''
  \emph{arXiv preprint}, vol. arXiv:1606.08415, 2016.

\bibitem{conneau2019unsupervised}
A.~Conneau, K.~Khandelwal, N.~Goyal, V.~Chaudhary, G.~Wenzek, F.~Guzm{\'a}n,
  E.~Grave, M.~Ott, L.~Zettlemoyer, and V.~Stoyanov, ``{Unsupervised
  Cross-lingual Representation Learning at Scale},'' in \emph{ACL}, 2020.

\bibitem{KingmaB14}
D.~P. Kingma and J.~Ba, ``{Adam: {A} Method for Stochastic Optimization},'' in
  \emph{ICLR}, 2015.

\bibitem{9023886}
V.-H. Nguyen, T.-C. Nguyen, M.-T. Nguyen, and N.~X. Hoai, ``{VNDS: A Vietnamese
  Dataset for Summarization},'' in \emph{NICS}, 2019.

\bibitem{lin-2004-rouge}
C.-Y. Lin, ``{{ROUGE}: A Package for Automatic Evaluation of Summaries},'' in
  \emph{Text Summarization Branches Out}, 2004.

\bibitem{tundik18_interspeech}
M.~Ákos Tündik, G.~Szaszák, G.~Gosztolya, and A.~Beke, ``{User-centric
  Evaluation of Automatic Punctuation in ASR Closed Captioning},'' in
  \emph{INTERSPEECH}, 2018.

\bibitem{NguyenNTDM20}
T.~B. Nguyen, Q.~M. Nguyen, H.~N.~T. Thu, Q.~T. Do, and L.~C. Mai, ``{Improving
  Vietnamese Named Entity Recognition from Speech Using Word Capitalization and
  Punctuation Recovery Models},'' in \emph{INTERSPEECH}, 2020.

\bibitem{huang21g_interspeech}
Q.~Huang, T.~Ko, H.~L. Tang, X.~Liu, and B.~Wu, ``{Token-Level Supervised
  Contrastive Learning for Punctuation Restoration},'' in \emph{INTERSPEECH},
  2021.

\bibitem{shi21_interspeech}
N.~Shi, W.~Wang, B.~Wang, J.~Li, X.~Liu, and Z.~Lin, ``{Incorporating External
  POS Tagger for Punctuation Restoration},'' in \emph{INTERSPEECH}, 2021.

\bibitem{NGUYEN20212020BDP0005}
T.~T.~H. NGUYEN \emph{et~al.}, ``{Toward Human-Friendly ASR Systems: Recovering
  Capitalization and Punctuation for Vietnamese Text},'' \emph{IEICE
  Transactions on Information and Systems}, no.~8, 2021.

\bibitem{capu_binhnguyen}
B.~Nguyen, V.~Nguyen, H.~Nguyen, P.~Pham, T.~L. Nguyen, T.~Do, and C.~Luong,
  ``{Fast and Accurate Capitalization and Punctuation for Automatic Speech
  Recognition Using Transformer and Chunk Merging},'' in \emph{O-COCOSDA},
  2019.

\bibitem{phost}
L.~T. Nguyen, N.~L. Tran, L.~Doan, M.~Luong, and D.~Q. Nguyen, ``{A
  High-Quality and Large-Scale Dataset for English-Vietnamese Speech
  Translation},'' in \emph{INTERSPEECH}, 2022, p. to appear.

\bibitem{rdrsegmenter}
D.~Q. Nguyen, D.~Q. Nguyen, T.~Vu, M.~Dras, and M.~Johnson, ``{A Fast and
  Accurate Vietnamese Word Segmenter},'' in \emph{LREC}, 2018.

\bibitem{vu-etal-2018-vncorenlp}
T.~Vu, D.~Q. Nguyen, D.~Q. Nguyen, M.~Dras, and M.~Johnson, ``{{V}n{C}ore{NLP}:
  A {V}ietnamese Natural Language Processing Toolkit},'' in \emph{NAACL
  (Demonstrations)}, 2018.

\end{thebibliography}
\bibliographystyle{IEEEtran}

\end{document}